\documentclass[10pt,twocolumn,letterpaper]{article}

\usepackage{wacv}                        

\usepackage{colortbl}   
\usepackage{makecell}   
\usepackage{multirow}
\usepackage{amssymb}
\usepackage{dsfont}     
\usepackage{pifont}

\providecommand{\mathbbm}[1]{\mathds{#1}}

\newcommand{\para}[2]{\smallskip\noindent\textbf{#1} #2}

\newcommand{\cmark}{\ding{51}}  
\newcommand{\xmark}{\ding{55}}  

\newcommand{\gtick}{\textcolor{green!60!black}{\cmark}}  
\newcommand{\rcross}{\textcolor{red}{\xmark}}            

\definecolor{best}{RGB}{255,200,100}
\definecolor{second}{RGB}{255,243,194}
\definecolor{lightgraytext}{gray}{0.55} 

\providecommand{\keywords}[1]{}

\AtBeginDocument{%
  \setlength{\abovedisplayskip}{5pt plus 2pt minus 2pt}%
  \setlength{\belowdisplayskip}{5pt plus 2pt minus 2pt}%
  \setlength{\abovedisplayshortskip}{2pt plus 1pt}%
  \setlength{\belowdisplayshortskip}{3pt plus 1pt}%
}

\definecolor{wacvblue}{rgb}{0.21,0.49,0.74}
\usepackage[pagebackref,breaklinks,colorlinks,allcolors=wacvblue]{hyperref}

\def\wacvPaperID{2342} 
\def\confName{WACV}
\def\confYear{2027}

\title{DecoGS: Adaptive Static-Dynamic \underline{Deco}upling of 3D \underline{G}aussians for \\ Free-Viewpoint Video \underline{S}treaming}

\author{Idil Sulo\\
University of Bonn, Almetra\\
\and
Alexey Supikov\\
V3DEO\\
\and
Ilke Demir\\
Cauth AI\\
\and
Sainan Liu\\
Intel Labs\\
}

\begin{document}

\twocolumn[{%
\renewcommand\twocolumn[1][]{#1}%
\maketitle
\begin{center}
    \vspace{-3px}
    \centering
    \captionsetup{type=figure}
    \includegraphics[width=.85\linewidth]{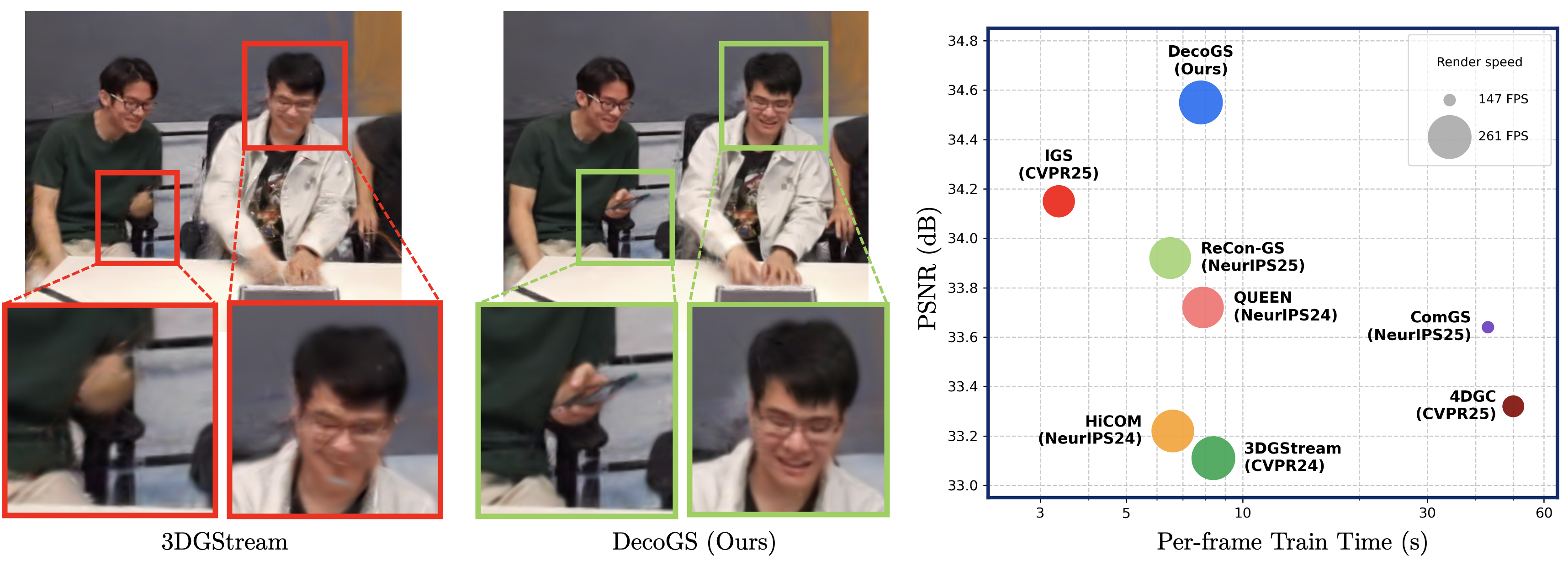}
    \captionof{figure}{\textbf{Quality and efficiency comparison.} DecoGS achieves superior reconstruction quality while maintaining efficient streaming performance. The left images illustrate high-fidelity rendering results produced by DecoGS in dynamic regions. The right plot compares DecoGS with prior streaming methods \cite{yan2025instant, gao2024hicom, sun20243dgstream, girish2024queen, fu2025recon, chen2025motion, hu20254dgc} demonstrating the effectiveness of static--dynamic decoupling.}
    \label{fig:teaser}
\end{center}%
}]

\begin{abstract}
Streaming 3D reconstruction demands both speed and temporal fidelity, goals that existing methods undermine by updating every Gaussian every frame, even in static regions. We present DecoGS, a method for efficient online training of 3D Gaussians from streaming videos. Unlike prior methods that update the entire scene indiscriminately, DecoGS introduces an adaptive mechanism that selectively focuses optimization on spatiotemporal regions exhibiting motion or photometric changes. This targeted training strategy eliminates redundant updates that cause flickering and drift in nominally static regions, while enabling fast, high-fidelity scene updates. The pipeline further integrates region-aware Gaussian management through gradient gating and efficient visibility filtering to maintain temporal coherence and a compact memory footprint. On N3DV and MeetRoom, DecoGS achieves 34.55 and 31.60 dB PSNR respectively, outperforming all streaming and offline baselines, while rendering at 261 FPS with $70\times$ lower temporal flicker than \vspace{-2px}the best prior method, requiring no large-scale pretraining.

\keywords{3D Gaussian Splatting \and Streaming Free-Viewpoint Videos}

\end{abstract}

\section{Introduction}
\label{sec:intro}
Free-Viewpoint Video (FVV) reconstruction from multi-view captures represents a fundamental challenge in immersive media, powering emerging applications in VR and AR. By enabling interactive and photorealistic exploration of dynamic scenes, FVV transcends traditional 2D video limitations, offering unprecedented viewer agency. Streaming-based FVV, with applications in remote telepresence and sports broadcasting, delivers such responsiveness at scale. This necessitates online 3D reconstruction that incrementally refines scene representations as new frames arrive, with a minimal latency in comparison to offline optimization, making it suitable for interactive applications.

The emergence of Neural Radiance Fields (NeRFs)~\cite{mildenhall2021nerf} enabled photorealistic scene reconstruction from multi-view images. Following this, synthesizing novel views gained significant popularity and several NeRF-based works~\cite{fang2022fast, li2022streamingmeet, li2022n3dv, li2021neural, li2023dynibar, park2021nerfies, park2021hypernerf, park2023temporal, pumarola2021d, wang2023neural, fridovich2023kplanes, xian2021space} attempted reconstruction of dynamic scenes for FVVs. However, most are limited by either requiring the full video before training (offline) or by failing to maintain temporal consistency as frames arrive (online), reducing utility for streaming.

Recent 3D Gaussian Splatting (3DGS)~\cite{kerbl3Dgaussians} methods further improve novel view synthesis with real-time rendering. However, adapting these \textit{static} techniques to streaming video remains challenging as new frames continuously reveal scene changes, requiring adaptation that is both fast and temporally consistent. Building on 3DGS,  \textit{offline training} methods \cite{yan20244dgaussians, li2024spacetime, wu20244dgaussians, yang2023real} demonstrate high-quality reconstruction from pre-recorded sequences. However, these methods require frames at all time steps to be present prior to training and scene reconstruction. Therefore, they are not practical for live streaming scenarios in which frames arrive sequentially over time. Recently, \cite{wang2025degauss, wu2025swift4d} explore separating static and dynamic scene components for dynamic reconstruction. However, these approaches operate in offline training settings that assume access to the full sequence and therefore do not address incremental optimization for streaming scenarios. Current \textit{online training} methods try to tackle these challenges in streaming scenarios. IGS~\cite{yan2025instant} reduces per-frame latency by pretraining large motion networks that infer Gaussian updates directly from video streams. While reducing per-frame processing time, it requires 192 GPU-hours of pretraining, transferring the computational burden rather than eliminating it. Another line of work \cite{sun20243dgstream, hu20254dgc} performs per-scene optimization by incrementally updating Gaussian parameters as new frames arrive. While effective for local adaptation, they re-optimize all Gaussians, causing gradient accumulation in static regions that manifests as temporal artifacts. 

We observe that in dynamic scenes, more than half of Gaussians do not require updates. Exploiting this sparsity is the key idea behind DecoGS: an incremental streaming framework that restricts optimization to dynamic regions. DecoGS provides an adaptive mechanism that estimates moving and deforming regions per frame, and restricts optimization to Gaussians within these regions while freezing gradient flow in stationary areas. This design suppresses redundant updates and concentrates optimization on the most informative regions, allowing DecoGS to train effectively while updating only as few as 35\% of the Gaussians per iteration. Combined with region-aware Gaussian management, adaptive update scheduling, and efficient visibility filtering, DecoGS produces compact and temporally coherent representations that enable real-time rendering and scalable deployment, while improving reconstruction quality.
In summary, our contributions include
\begin{itemize}
    \item a real-time streaming 3D reconstruction method that achieves superior quality with efficient online training,
    \item an adaptive static-dynamic decoupling module that identifies and freezes stable scene regions,
    \item a dynamic selection framework that limits the number of updated Gaussians per iteration, 
    preserving temporal coherence via gradient gating and visibility filtering.
\end{itemize}

Experiments on two real-world datasets across 9 dynamic scenes validate that DecoGS's reconstruction quality of 34.55 (N3DV) and 31.60  (MeetRoom) PSNR surpasses SOTA online \emph{and} offline methods, also providing the fastest rendering with 260-261 FPS. Furthermore, we quantify the temporal coherence visible in our supplemental videos with 0.003/0.004 mTV on two scenes, +0.079/+0.099 improvement over the second best approach. DecoGS establishes a new direction for practical and resource-efficient 3D Gaussian learning from continuous video streams, with a competitive advantage in rendering quality and training time.

\section{Related Work}
\label{sec:related_work}

\subsection{Novel View Synthesis for Static Scenes}

Synthesizing novel views of static scenes has long been a fundamental topic in computer vision and computer graphics, with early approaches ~\cite{buehler2001unstructured,chai2000plenoptic,davis2012unstructured,gortler2023lumigraph,levoy2023light,shum1999rendering} achieving novel-view generation through view interpolation and image-based warping. With the advent of deep learning, Neural Radiance Fields (NeRF)~\cite{mildenhall2021nerf} revolutionized the field by representing continuous scene radiance via coordinate-based neural networks, producing photorealistic novel views from sparse input images.
A wide range of subsequent methods improved NeRF in different dimensions by accelerating training \cite{chen2022tensorf, chen2023dictionary, fridovich2022plenoxels, hu2023tri, muller2022instant, sun2022direct}, enabling real-time rendering \cite{chen2023mobilenerf, garbin2021fastnerf, hedman2021baking, reiser2023merf, wizadwongsa2021nex, yu2021plenoctrees}, and enhancing reconstruction quality for complex or sparse scenes \cite{barron2021mip, barron2022mip, barron2023zip, martin2021nerf, mildenhall2022nerf, park2023temporal, wimbauer2023behind, wynn2023diffusionerf, yang2023freenerf, yu2021pixelnerf, verbin2024ref, chen2021mvsnerf}. However, the original volumetric formulation of NeRF requires dense sampling and neural network evaluation along each camera ray, which introduces inherent trade-offs among speed, quality, and storage efficiency. To address these constraints, 3D Gaussian Splatting (3DGS)~\cite{kerbl3Dgaussians} introduced a point-based representation in which each scene element is modeled as a Gaussian primitive with learnable opacity, scale, and spherical harmonic color coefficients, enabling real-time, high-fidelity view synthesis. Building on this foundation, our work leverages 3D Gaussian representations not only for static view synthesis but also for efficient online training on continuous video streams, enabling dynamic Free-Viewpoint Video (FVV) reconstruction with minimal computational overhead.

\subsection{Free-Viewpoint Videos for Dynamic Scenes}
Extending novel view synthesis from static to dynamic scenes has become an active area of research, with early progress driven by NeRF-based dynamic representations \cite{attal2023hyperreel, cao2023hexplane, fang2022fast, li2022streamingmeet, li2022tava, li2022n3dv, li2021neural, li2023dynibar, park2021nerfies, park2021hypernerf, park2023temporal, pumarola2021dnerf, fridovich2023kplanes, song2023nerfplayer, tretschk2021non, wang2023mixed, wang2023neural, wang2023tracking, weng2022humannerf, zhao2022humannerf, yang2022banmo, jiang2024robust, jiang2024hifi4g}. These approaches model motion by conditioning neural fields on time or deformation networks, but remain limited by slow optimization and heavy volumetric rendering. Following 3DGS\cite{kerbl3Dgaussians}, several works have explored dynamic Gaussian formulations \cite{huang2024sc, li2024spacetime, wu20244dgaussians, yan20244dgaussians, yang2024deformable, yang2023real, kratimenos2024dynmf} integrating 3DGS's real-time capabilities into temporal modeling frameworks. Recently, \cite{wang2025degauss, wu2025localdygs, wu2025swift4d, he2024s4d, yan20244dgaussians, liang2025gaufre} attempt to separate dynamic and static Gaussian points and introduce external models to segment foreground and background areas. ClipGStream~\cite{liang2026clipgstream} further scales offline reconstruction to long sequences and large-scale motion via clip-level optimization. Although these methods achieve high-fidelity reconstruction, they typically rely on offline optimization over full video sequences, making them unsuitable for applications demanding fast updates and low latency, such as FVV and immersive VR/AR streaming.

To overcome this limitation, streaming-based approaches have been proposed. Among these, earlier methods such as StreamRF~\cite{li2022streamingmeet}, NeRFPlayer~\cite{song2023nerfplayer} and ReRF~\cite{wang2023neuralrerf} utilize NeRF-based dynamic representations, while recent streaming approaches \cite{sun20243dgstream, hu20254dgc, yan2025instant} utilize Gaussian representations. These methods formulate dynamic reconstruction as an incremental learning problem, updating the scene representation as new frames arrive. In particular, 3DGStream \cite{sun20243dgstream} models Gaussian motion between frames, significantly improving rendering speed and memory efficiency; however, it fails to represent dynamic areas with high quality. Several works \cite{hu20254dgc, gao2024hicom, chen2025motion, fu2025recon, girish2024queen} model the motion in a similar fashion, focusing on compactness to reduce storage cost. Among these, QUEEN~\cite{girish2024queen} improves streaming efficiency by encoding temporal Gaussian updates with quantized residuals and sparse position updates, focusing on compression of changes rather than explicitly separating static and dynamic Gaussians during optimization. Furthermore, these methods still perform per-frame optimization over all Gaussians, leading to substantial computational overhead and artifacts arising from updates in static regions. Recent variants, such as IGS~\cite{yan2025instant}, aim to further reduce per-frame latency and improve reconstruction quality by pretraining a motion network to predict Gaussian deformation directly from streaming video. While effective, this approach requires extensive offline training prior to incremental learning, and falls short in generalization and quality of fine details. In contrast, DecoGS introduces a lightweight, adaptive formulation that updates only the dynamic areas of the scene, heavily suppressing the emergence of artifacts and improving temporal consistency.

\section{Preliminary}
\label{sec:preliminary}
\textbf{3D Gaussian Splatting (3DGS)} \cite{kerbl3Dgaussians} represents static scenes as a collection of anisotropic 3D Gaussians (3DGs), where the color of each pixel is obtained through point-based alpha blending. Each 3DG is parameterized by a center $\boldsymbol{\mu} \in \mathbb{R}^3$ and a covariance matrix $\Sigma \in \mathbb{R}^{3 \times 3}$. For every 3D point $x \in \mathbb{R}^{3}$, the 3DG is defined as:

\begin{equation}
G(x ; \boldsymbol{\mu}, \Sigma)=e^{-\frac{1}{2}(x-\boldsymbol{\mu})^T \Sigma^{-1}(x-\boldsymbol{\mu})},
\end{equation}

\noindent where covariance matrix $\Sigma$ is given by, with scale matrix $\mathbf{s}$ and rotation matrix $\mathbf{R}$:

\begin{equation}
\Sigma=\mathbf{R} \operatorname{diag}(\mathbf{s}) \operatorname{diag}(\mathbf{s})^T \mathbf{R}^T.
\end{equation}

For rendering, the 3DG is projected onto the 2D space and the Gaussians covering a pixel are sorted based on their depth. The color of the pixel $\mathbf{c}$ is obtained using point-based alpha blending using the color of the $i$-th Gaussian, $c_i$, opacity value $\alpha_i$ computed after particle projection:

\begin{equation}
\mathbf{c}=\sum_{i=1}^n c_i \alpha_i \prod_{j=1}^{i-1}\left(1-\alpha_j\right),
\end{equation}

\textbf{Neural Transformation Cache (NTC).} Following \cite{sun20243dgstream}, we employ NTC, a lightweight MLP-based network that predicts per-Gaussian transformation changes to capture scene dynamics. At each timestep, NTC estimates spatial and rotational residuals $(d_{rot},d\mu)$ conditioned on the Gaussian’s latent attributes, enabling rapid adaptation of positions and orientations without full retraining. Unlike directly optimizing all 3DGs, NTC operates as a compact motion prior that is queried and updated online, substantially reducing computational overhead. Its outputs are later committed back to the explicit Gaussian parameters during densification, forming a bridge between neural field prediction and explicit 3D representation refinement.

\begin{figure*}[t] 
  \centering
  \includegraphics[width=.84\linewidth]{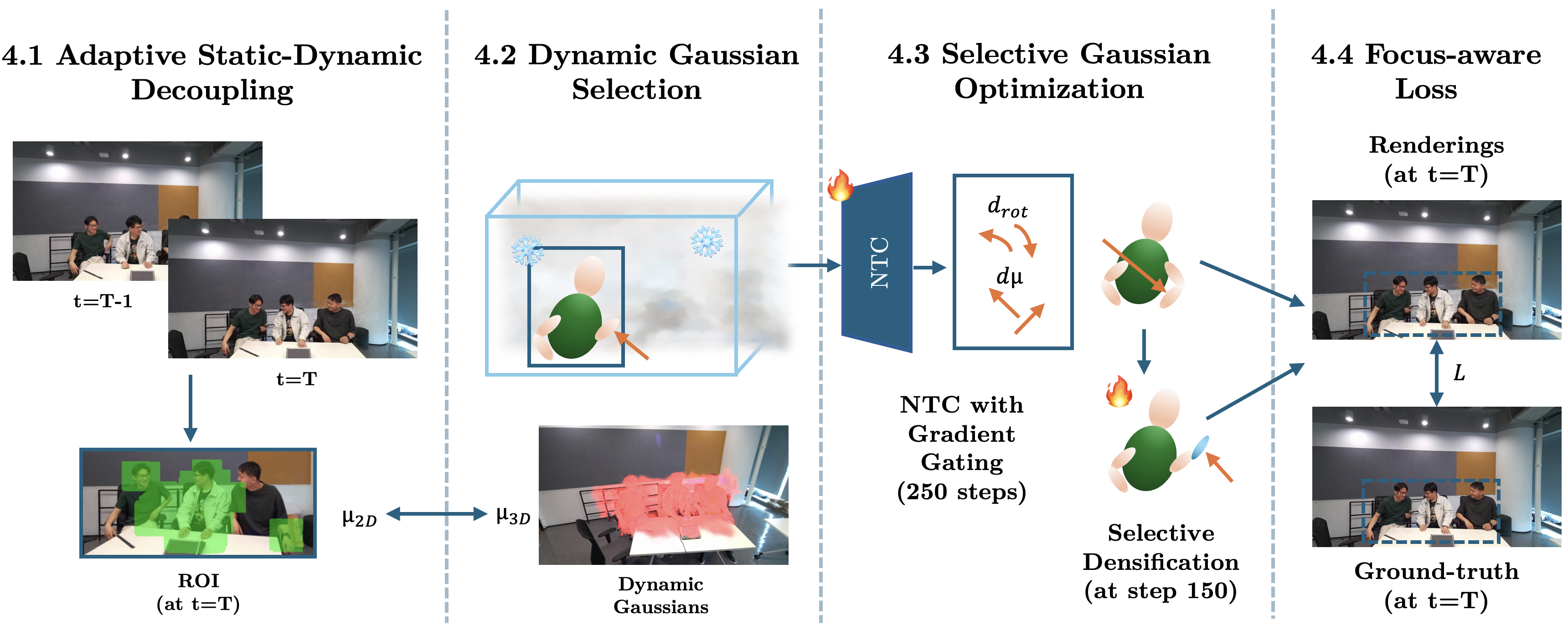}
  \caption{\textbf{Overview of DecoGS.} Given a set of multi-view video streams, DecoGS reconstructs an FVV stream of the dynamic scene. At $t=0$, we optimize a set of initial 3D Gaussians, and for each $t > 0$, we compute a pixel-wise difference mask to isolate dynamic regions, dilated via max-pooling to capture motion boundaries. Throughout training, we optimize the NTC for 250 steps to rotate and translate Gaussians from the previous timestep $t=T-1$. At step 150, we employ selective densification that limits the spawning of Gaussians only to the selected areas, enabling the emergence of new objects. From this step onward, we optimize the selected Gaussians jointly with NTC using a focus-aware loss that concentrates the optimization to the detected ROI.}
  \label{fig:overview}
  \vspace{-8pt} 
\end{figure*}

\section{Method}
\label{sec:method}

Given multi-view video streams as input, DecoGS constructs photo-realistic FVV streams via online training of an incremental Gaussian Splatting framework. We begin by training a static 3DGS model at the initial time step $t=0$. For each subsequent timestep $t>0$, we apply an efficient Adaptive Static-Dynamic Decoupling strategy (\cref{sec:static-dynamic-decoupling}) to isolate motion regions in 2D image space. Using this information, we perform Dynamic Gaussian Selection (\cref{sec:dynamic-gaussian-selection}), which identifies the subset of Gaussians affected by dynamic content. The selected subset is then refined through Selective Gaussian Optimization (\cref{sec:selective-gaussian-optimization}), while all static Gaussians remain frozen to preserve stability and reduce redundancy. Finally, we describe a Focus-Aware Loss Function (\cref{sec:focus-aware-loss-func}) that emphasizes dynamic areas during optimization, further improving temporal coherence and visual sharpness. Overview is provided in \cref{fig:overview}.

\subsection{Adaptive Static-Dynamic Decoupling}
\label{sec:static-dynamic-decoupling}
At each time step $t$, we receive a new set of multi-view frames $\{I_t^v\}$ from cameras $v \in \{1, \ldots, V\}$.  
We first compute a \textit{difference mask} $\tilde{M}_t^v$ in the pixel space between consecutive frames $I_{t-1}^v$ and $I_t^v$ to identify dynamic regions.  
This step identifies photometrically changed regions without optical flow or learned features, relying instead on motion and appearance changes being well-approximated by per-pixel intensity differences.

Formally, $\tilde{M}_t^v$ is computed for each view $v$ as follows:
\begin{equation}
    D_t^v(x,y) = \max_{i_c} |I_t^v(x,y,i_c) - I_{t-1}^v(x,y,i_c)|,
\end{equation}
\begin{equation}
    M_t^v(x,y) = \mathbbm{1} [D_t^v(x,y) > \tau],
\end{equation}
where $\tau$ is an intensity threshold, $i_c$ is a color channel. To account for motion boundaries and slight localization errors, we spatially dilate $M_t^v$ via a max-pooling operation:
\begin{equation}
    \tilde{M}_t^v = \mathrm{MaxPool}(M_t^v,\, r),
\end{equation}
with kernel radius $r$, yielding a pixel-wise selection mask used to identify dynamic Gaussians.

\subsection{Dynamic Gaussian Selection}
After detecting ROIs in 2D, we need to associate these regions with 3D Gaussians of interest. To seek a low-cost and efficient streaming procedure that does not increase the overall computation, we use camera parameters to project Gaussian centers from 3D world space to 2D image space.

\label{sec:dynamic-gaussian-selection}
\para{Projection of 3D Gaussians.}{To associate Gaussians with image-space changes, we project 3D centers $\boldsymbol{\mu}_i^{3D} \in \mathbb{R}^3$ into camera coordinates via calibrated intrinsics/extrinsics:}
\begin{equation}
\label{eq:projection}
\begin{aligned}
\boldsymbol{\mu}_i^{2D,v} &= \Pi_v(\boldsymbol{\mu}_i^{3D}) =
\begin{bmatrix}
u_i^v \\ v_i^v
\end{bmatrix}
=
\begin{bmatrix}
f_x \frac{X_i^v}{Z_i^v} + c_x \\
f_y \frac{Y_i^v}{Z_i^v} + c_y
\end{bmatrix},
\end{aligned}
\end{equation}
where $(X_i^v, Y_i^v, Z_i^v)$ are the coordinates of the Gaussian in camera space, and $(f_x, f_y, c_x, c_y)$ denote the intrinsic parameters. This projection efficiently identifies visible Gaussians with positive depth and avoids costly procedures such as 2D-to-3D motion feature lifting.

\para{Mask-based Gaussian Selection.}{Given the per-view dilated masks $\tilde{M}_t^v$ from \cref{sec:static-dynamic-decoupling}, we select Gaussians whose projected centers fall inside at least $\kappa$ views’ active pixels. Let $\mathcal{S}_t$ denote the selected set:}
\begin{equation}
    \label{eq:selection}
    \mathcal{S}_t
    = \bigl\{\, i \,\big|\,
    \sum_{v=1}^{V} \mathbbm{1}\!\bigl[\tilde{M}_t^v\!\bigl(\boldsymbol{\mu}_i^{\mathrm{2D},v}\bigr) = 1\bigr]
    \ge \kappa \,\bigr\}.
\end{equation}

In practice, we set $\kappa=2$. Additionally, we intersect this selection with per-view visibility filters to suppress updates to occluded Gaussians, yielding the final \textit{global Gaussian mask}. Gradients for 3DGs outside this mask are blocked: \begin{equation} \frac{\partial \mathcal{L}}{\partial \theta_i} = 0, \quad \text{for } i \notin \mathcal{S}_t. \end{equation}

\subsection{Selective Gaussian Optimization}
\label{sec:selective-gaussian-optimization}

\para{NTC with Gradient Gating.}{Following \cite{muller2022instant, sun20243dgstream} due to its compactness and efficiency, we extend NTC with a gradient gating mechanism. During incremental training, the model retains all Gaussians for rendering but only backpropagates gradients to $\mathcal{S}_t$. This targeted optimization suppresses redundant updates in static areas while preserving details in dynamic ones. For each iteration:}
\begin{equation}
\mathbf{\theta}_i \leftarrow
\begin{cases}
\mathbf{\theta}_i - \eta \, \nabla_{\mathbf{\theta}_i} \mathcal{L}, & \text{if } i \in \mathcal{S}_t, \\[3pt]
\mathbf{\theta}_i, & \text{otherwise,}
\end{cases}
\end{equation}
where $\eta$ is the learning rate. This gating is implemented efficiently as an in-place tensor operation, avoiding overhead in both forward and backward passes. We observe that the gradient gating mechanism significantly reduces updates to 3D Gaussians in static regions, such as the walls and ceiling, while concentrating training updates in dynamic regions.

\para{Optimizer State Reset.} Gradient gating alone is insufficient to prevent drift: even when parameter updates are blocked, Adam's first and second moment buffers $m_i, v_i$ for static Gaussians continue to accumulate stale signal from dynamic frames, corrupting future optimization steps when those Gaussians later enter $\mathcal{S}_t$. We therefore zero these buffers after every update for all Gaussians outside $\mathcal{S}_t$:

\begin{equation}
    m_i \leftarrow 0, \quad v_i \leftarrow 0, \qquad \forall\, i \notin \mathcal{S}_t,
\end{equation}
This prohibits gradient history from dynamic frames to corrupt the optimization of static Gaussians in future steps.

\para{Selective Densification.}{The selective transformation of Gaussians covers a large portion of dynamics in the scenes by managing occlusions and disappearances in subsequent timesteps. Still, this approach falls short in capturing the emergence of new objects, such as a smartphone being taken out of a pocket. It is not feasible to generate an extensive number of additional Gaussians, as this would create intractable growth. Therefore, a reliable strategy for estimating the emergence of new Gaussians is required.} 

\cite{sun20243dgstream} proposes a mechanism to capture every potential location where new objects might emerge by tracking the view-space positional gradient during the training of the NTC. We extend this by adding a selective densification process that restricts the spawned Gaussians to the ROI. First, we prune and clone only within the ROI. This limits the substantial growth of Gaussians, spawned only in the required regions. Empirically, Gaussians selected for densification never exceed 35\% of the dynamic scene. For high-gradient regions where new 3D Gaussians are spawned, we set a lower gradient threshold of $\tau_d=0.0001$.
Gradient and radius accumulators used by 3DGS densification are zeroed outside. To prevent off-ROI drift, any added Gaussian whose projections fall outside all per-view ROIs is repositioned to its parent's location before being committed to the scene. Finally, we perform a mask re-sync. Since the number of points changes after the prune-and-clone step, the external mask is resized to match the new set of 3D Gaussians.

\subsection{Focus-aware Loss Function}
\label{sec:focus-aware-loss-func}
Although the centers of the selected Gaussians lie within the ROI, updating the Gaussian can affect regions outside these boundaries, depending on its size. Therefore, we do not strictly limit the optimization to this region. Instead, to emphasize dynamic regions, we introduce a focus-weighted reconstruction loss that gives higher weight to pixels within the ROI.  
The overall loss combines a global $\ell_1$ term, a focus-weighted $\ell_1$ term, and a focus-only SSIM term:
\begin{align}
\label{eq:loss}
\mathcal{L} &= (1-\lambda_{\text{ssim}})\,[\beta_1\,\mathcal{L}_{1} + \beta_2\,\mathcal{L}_{1}^{\text{focus}}] + \lambda_{\text{ssim}}\,\mathcal{L}_{\text{ssim}}^{\text{focus}},\\
\mathcal{L}_{1}^{\text{focus}} &= \|\tilde{M}_t^v \odot (I_t^v - \hat{I}_t^v)\|_1, \\
\mathcal{L}_{\text{ssim}}^{\text{focus}} &= 1 - \mathrm{SSIM}(\tilde{M}_t^v \odot (I_t^v - \hat{I}_t^v)),
\end{align}
and $\beta_1$, $\beta_2$ balance $\ell_1$ terms. SSIM is applied exclusively to the focus region, since gradients outside the ROI are already suppressed by the gradient-gating mechanism.

\section{Experiments}
\label{sec:experiments}

\subsection{Setup}
\para{Datasets.}{We perform our experiments on two real-world dynamic scene datasets, \ie N3DV dataset \cite{li2022n3dv} and MeetRoom dataset \cite{li2022streamingmeet}. Each dataset reserves 1 camera view as a held-out test set, while the remaining views are used for training. \textbf{N3DV dataset} \cite{li2022n3dv} is captured via 21 multi-view cameras and includes dynamic scenes recorded at a resolution of $2704 \times 2028$ at 30 FPS. For a fair comparison with methods that pre-train on sequences from the same dataset, we report results on the N3DV test sequences over 300 frames. 
The \textbf{MeetRoom dataset} \cite{li2022streamingmeet} is captured via 13 multi-view Azure Kinect cameras and includes dynamic scenes recorded at a resolution of $1280 \times 720$ and 30 FPS. This dataset presents more challenging scenarios with rapid changes and motion blur.}

\para{Implementation Details.}{We utilize 3DGS \cite{kerbl3Dgaussians} and NTC using InstantNGP \cite{muller2022instant, tiny-cuda-nn}. For the training of the initial 3DGS, we adapt the learning rates from the N3DV dataset defaults and use the same parameters for the MeetRoom dataset. To suppress any noise that might occur during the transformation of the initial static areas, we train the NTC at 500 iterations only for the initial frame. For the remainder of the video sequence, we train NTC for 250 iterations and limit the Gaussian optimization to the last 100 iterations after selective densification. Gaussians corresponding to static areas remain unchanged throughout training. Therefore, we perform a full training of the original 3DGS. All experiments are conducted on an NVIDIA RTX 4090 GPU. We provide further details in Supp. Sec. B.}

\begin{table*}[t]
    \centering
    \begin{minipage}[t]{0.5\linewidth}
        \centering
        \caption{\textbf{Comparison on N3DV} of offline and online 3DGS methods at $1352 \times 1014$. \colorbox{best}{Best} and \colorbox{second}{second best} are highlighted. Pre-train-free (\gtick{}/\rcross{}) indicates whether a method operates without any pre-training beyond initialization. ($\dagger$) marks results reproduced by us with the official code in the same experimental environment. }
        \label{tab:n3dv}
        \setlength{\tabcolsep}{2pt}
        \small
        \resizebox{\linewidth}{!}{%
        \begin{tabular}{lcccccc}
        \toprule
        Method & PSNR $\uparrow$ & PSNR$\uparrow$ & SSIM$\uparrow$ & Render$\uparrow$ & Pre-train-free & Per-frame$\downarrow$\\
        & Test (dB) & (dB) & & (FPS) & (\gtick{}/\rcross{}) & Train (s) \\
        \midrule
        Offline Training & & & & & \\
        \midrule
        K-Planes \cite{fridovich2023kplanes}     & 32.17 & 31.63 & - & 0.15 & \rcross & -\\
        Realtime-4DGS \cite{yang2023real} & 33.94 & 32.01 & \cellcolor{best}0.972 & 114 & \rcross & - \\
        4DGS \cite{wu20244dgaussians}         & 32.70 & 31.15 & 0.950 & 30 & \rcross & 7.8 \\
        SpaceTime-GS \cite{li2024spacetime}  & 33.71  & 32.05 & \cellcolor{best}0.972 & 140 & \rcross & - \\
        Saro-GS \cite{yan20244dgaussians}      & 33.90  & 32.15 & - & 40 & \rcross & - \\
        Swift4D \cite{wu2025swift4d} & - & 32.23 & \cellcolor{best}0.972 & 125 & \rcross & \cellcolor{second}5 \\
        \midrule
        \midrule
        Online Training & & & & & \\
        \midrule
        StreamRF \cite{li2022streamingmeet} & 32.09 & 30.68 & - & 8.3 & \gtick & 15\\
        4DGC \cite{hu20254dgc}   & 33.32 & 31.58 & - & 168 & \gtick & 50 \\
        IGS \cite{yan2025instant} & \cellcolor{second}34.15 & - & - & 204 & \rcross~(192 hours) & \cellcolor{best}3.35 \\
        HiCoM$^\dagger$ \cite{gao2024hicom}   & 33.22 & 32.08 & 0.953 & \cellcolor{second}255 & \gtick & 6.6 \\
        ReCon-GS \cite{fu2025recon}   & 33.92 & \cellcolor{second}32.66 & \cellcolor{second}0.956 & 250 & \gtick & 6.4 \\
        ComGS \cite{chen2025motion} & 33.64 & 32.12 & - & 147 & \gtick & 43 \\
        QUEEN \cite{girish2024queen} & 33.72 & 32.01 & 0.946 & 248 & \gtick & 7.9\\
        MoRGS \cite{lee2026morgs} & - & 32.53 & 0.950 & 200 & \gtick & - \\
        3DGStream$^\dagger$ \cite{sun20243dgstream}& 33.03 & 31.35 & 0.948 & \cellcolor{best}261 & \gtick & 8.4 \\
        Ours & \cellcolor{best}34.55 & \cellcolor{best}33.36& \cellcolor{second}0.956 & \cellcolor{best}261 & \gtick & 7.8 \\
        \bottomrule
        \end{tabular}
        }
    \end{minipage}
    \hfill
    \begin{minipage}[t]{0.48\linewidth}
        \centering
        \caption{\textbf{Comparison on MeetRoom} of streaming methods at $1280 \times 720 $. Train denotes seconds at each iteration during incremental training \colorbox{best}{Best} and \colorbox{second}{second best} are highlighted.}
        \label{tab:meetroom}
        \setlength{\tabcolsep}{3pt}
        \small
        \resizebox{\linewidth}{!}{%
        \begin{tabular}{lcccc}
        \toprule
        Method & \begin{tabular}{c}
        PSNR $\uparrow$ \\
        (dB)
        \end{tabular} & \begin{tabular}{c}
        SSIM $\uparrow$ \\
        \\
        \end{tabular} & \begin{tabular}{c}
        Train $\downarrow$ \\
        (s)
        \end{tabular} & \begin{tabular}{c}
        Render $\uparrow$ \\
        (FPS)
        \end{tabular} \\
        \midrule
        4DGC \cite{hu20254dgc} & 28.08 & - & 49.8 & 213 \\
        HiCoM$^\dagger$ \cite{gao2024hicom} & 29.57 & 0.944 & \cellcolor{second}3.9 & 236 \\
        ReCon-GS \cite{fu2025recon} & 30.84 & -  & \cellcolor{second}3.9 & \cellcolor{second}256 \\
        ComGS \cite{chen2025motion} & \cellcolor{second}31.49 & \cellcolor{best}0.955 & 28.3 & 98 \\
        3DGStream$^\dagger$ \cite{sun20243dgstream}  & 30.79 & \cellcolor{second}0.950 & 4.9 & \cellcolor{best}260 \\
        IGS \cite{yan2025instant} & 30.13 & - & \cellcolor{best}2.7 & 252 \\
        DecoGS (Ours) &	\cellcolor{best}31.60 & \cellcolor{best}0.955 & 4.3 & \cellcolor{best}260 \\
        \bottomrule
        \end{tabular}
        }

        \vspace{8pt}
        \caption{\textbf{Temporal consistency in static regions.} Following regions of \cite{yun2025compensating}, we report PSNR, SSIM and mTV (masked Total Variation) with standard deviations.}
        \label{tab:mtv}
        \setlength{\tabcolsep}{4pt}
        \small
        \resizebox{\linewidth}{!}{%
        \begin{tabular}{l|cc|cc}
        \multirow{2}{*}{Method} & \multicolumn{2}{c}{Coffee Martini} & \multicolumn{2}{c}{Flame Steak} \\
        \cmidrule(lr){2-3}\cmidrule(lr){4-5}
        & PSNR $\uparrow$ & $\text{mTV}_{\times 100}$ $\downarrow$  & PSNR $\uparrow$  & $\text{mTV}_{\times 100}$ $\downarrow$ \\
        \hline
        3DGStream & $27.75_{\pm 0.36}$ & $0.213_{\pm 0.040}$ & $33.47_{\pm 1.01}$ & $0.140_{\pm 0.026}$  \\
        3DGStream w. \cite{yun2025compensating} & \cellcolor{second}$29.54_{\pm 0.06}$ & \cellcolor{second}$0.082_{\pm 0.070}$ &  \cellcolor{second}$34.86_{\pm 0.37}$ & \cellcolor{second}$0.103_{\pm 0.046}$  \\
        DecoGS (Ours) & \cellcolor{best}$31.15_{\pm 0.24}$ & \cellcolor{best}$0.003_{\pm 0.003}$ & \cellcolor{best}$35.54_{\pm 1.03}$ & \cellcolor{best}$0.004_{\pm 0.013}$ \\
        \end{tabular}
        }
    \end{minipage}
\end{table*}

\subsection{Baselines}
We compare our approach to both state-of-the-art online and offline training methods for dynamic scene reconstruction. Offline methods \cite{yan20244dgaussians, li2024spacetime, wu20244dgaussians, yang2023real, fridovich2023kplanes, wu2025swift4d} use Gaussian primitives or hex-plane representations for the entire scene, but require the complete video sequence before training begins, an assumption incompatible with live streaming. In contrast, online methods \cite{li2022streamingmeet, hu20254dgc, yan2025instant, sun20243dgstream, girish2024queen, gao2024hicom, chen2025motion, fu2025recon, lee2026morgs} employ per-frame optimization, suitable for streaming by design. Among these, IGS \cite{yan2025instant} requires heavy pre-training prior to per-frame optimization on 4 scenes of N3DV. Therefore, we additionally provide mean PSNR across the test sequences ``cut beef'' and ``sear steak'' for a fair comparison. Similarly, we compare MoRGS \cite{lee2026morgs} only on N3DV via its self-reported results, as no public implementation is available (Supp. Sec. C.1).

\begin{figure*}[t]
    \centering
    \includegraphics[width=0.75\linewidth]{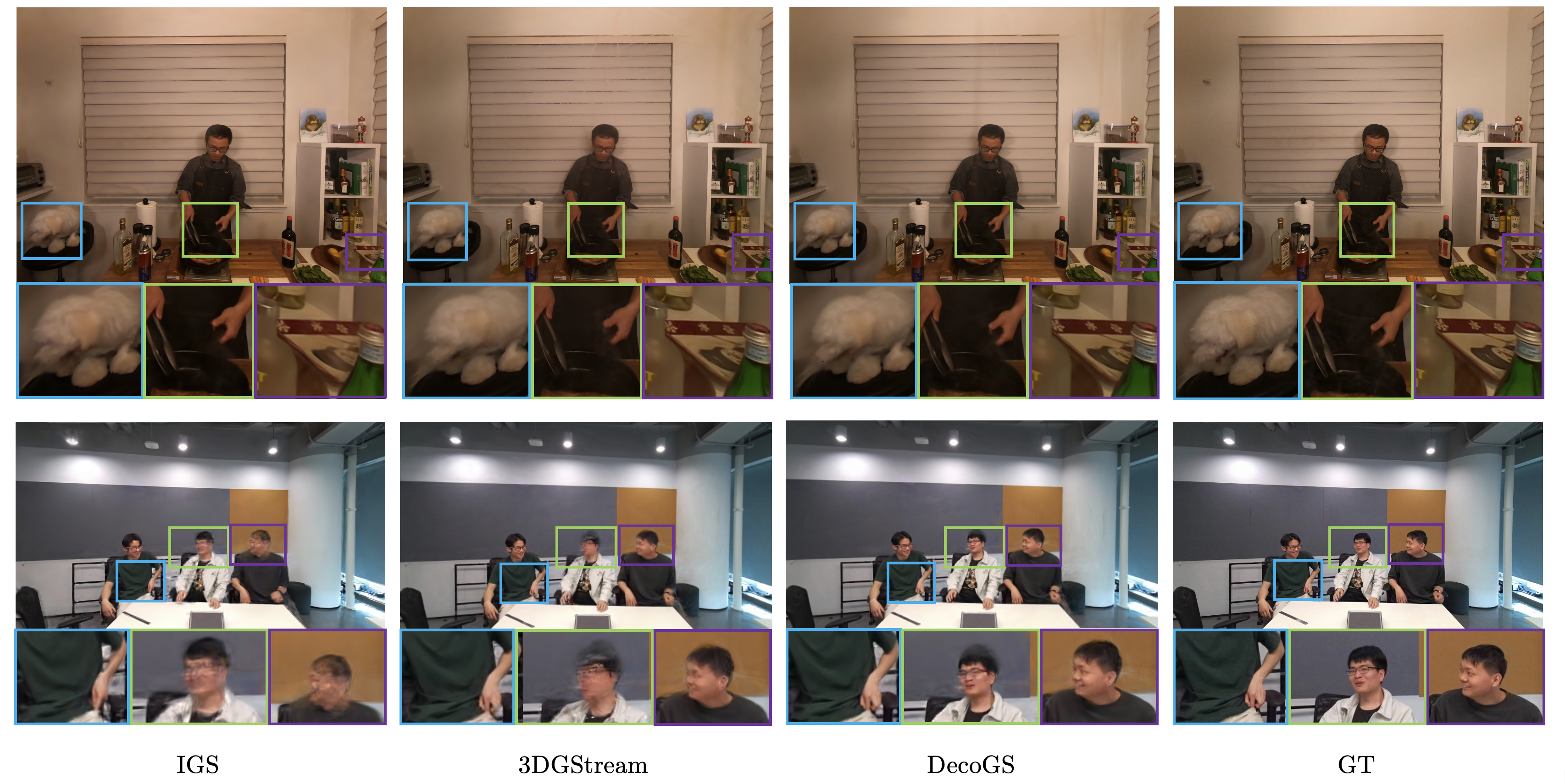}
    \caption{\textbf{Qualitative comparison} across streaming baselines IGS \cite{yan2025instant} and 3DGStream \cite{sun20243dgstream} on sear steak scene of N3DV dataset and the discussion scene of MeetRoom dataset. Our method achieves high-quality reconstruction both in static and dynamic regions.}
    \label{fig:qualitative}
\end{figure*}

\subsection{Comparisons}

\para{Quantitative Analysis.}{ We conduct quantitative comparisons to benchmark DecoGS on N3DV and MeetRoom datasets. In \cref{tab:n3dv}, we present peak signal-to-noise ratio (PSNR) for test and all sequences, rendering speed, whether pre-training is required prior to streaming, and per-frame training time. To highlight the superior quality of DecoGS, we compare it with both offline training methods that require the entire video sequence and online methods that target video streaming. For each scene, the metrics are computed as averages over 300 frames of the sequence. In addition, we provide per-scene results for all N3DV sequences in Supp. Sec. C.1.
While maintaining comparable rendering speed to 3DGStream, our method achieves higher rendering quality with a +2.01~dB increase in PSNR over all scenes. Notably, each baseline concentrates on a different point of the design space. IGS \cite{yan2025instant} attains the lowest per-frame training time by amortizing motion estimation into a large pre-trained network, trading it off against 192 GPU-hours of pre-training before streaming can start; this cost does not appear in DecoGS, a property shared by the remaining baselines. Despite requiring no pre-training, we still outperform IGS by +0.40~dB in PSNR over the two test sequences. Among recently introduced online methods, 4DGC~\cite{hu20254dgc}, HiCoM~\cite{gao2024hicom}, ComGS~\cite{chen2025motion}, and ReCon-GS~\cite{fu2025recon} mainly concentrate on compactness and storage, and therefore trade off either per-frame training time (50 and 43 seconds for 4DGC and ComGS) or reconstruction quality (33.22 and 33.92~dB for HiCoM and ReCon-GS). DecoGS instead concentrates on reconstruction quality and rendering speed, outperforming all of them on both PSNR metrics, as well as QUEEN~\cite{girish2024queen} and the flow-guided MoRGS~\cite{lee2026morgs}, while providing the fastest rendering at 261 FPS. Our method achieves state-of-the-art rendering quality for N3DV compared with both streaming and offline methods.}

To demonstrate the generalization of our method, we additionally conduct experiments on the MeetRoom dataset of StreamRF \cite{li2022streamingmeet} and report quantitative comparisons across FVV streaming baselines. As shown in \cref{tab:meetroom}, our method achieves state-of-the-art rendering quality compared to the baselines while maintaining fast online training and real-time rendering capabilities. Compared with 3DGStream, our method preserves on-par training time while achieving a +0.81~dB increase in PSNR, demonstrating our effectiveness in eliminating 3D Gaussians corresponding to static areas. Among the recent baselines, ComGS~\cite{chen2025motion} is the closest competitor at 31.49~dB; however, its focus on compactness comes at the cost of 28.3 seconds of per-frame training and 98 FPS rendering, whereas DecoGS surpasses it with 31.60~dB while training in 4.3 seconds per frame and rendering at 260 FPS. To validate our claims, we also report and compare PSNRs for the dynamic regions in Supp. Sec. C.1 for the MeetRoom scenes. 

Beyond global PSNR, we evaluate temporal stability in static regions using masked Total Variation (mTV)~\cite{yun2025compensating}, which measures per-pixel intensity change across consecutive frames. \cref{tab:mtv} reveals the most striking result: DecoGS achieves near-zero mTV ($0.003$ and $0.004$) on the ``coffee martini'' and ``flame steak'' scenes, reducing temporal flicker up to $\sim\!70\times$ over 3DGStream and outperforming the dedicated stabilization of~\cite{yun2025compensating} without any additional post-processing step. This near-zero flicker is a direct consequence of gradient gating: static Gaussians are never updated, so they cannot drift. Per-scene mTV comparisons across the full N3DV dataset are provided in Supp. Sec. C.1.

\para{Qualitative Analysis. }{We present qualitative comparisons across representative dynamic scenes from the N3DV and MeetRoom datasets. As in \cref{fig:qualitative}, our method consistently yields sharper and more temporally coherent reconstructions. In highly dynamic regions (hand motion, facial expression, and smoke from the steak) DecoGS captures fine appearance variations while keeping background consistency. Static objects, such as the snowman on the plate, remain artifact-free precisely because their Gaussians are never updated. In contrast, 3DGStream \cite{sun20243dgstream} exhibits noticeable ghosting and temporal flicker due to frame-wise optimization over the entire Gaussian set, while IGS \cite{yan2025instant} suffers from poor generalization on the discussion scene of MeetRoom due to pre-training on a limited subset of N3DV.}

\begin{figure*}[t]
    \centering
    \begin{minipage}[t]{0.41\linewidth}
        \centering
        \includegraphics[width=\linewidth]{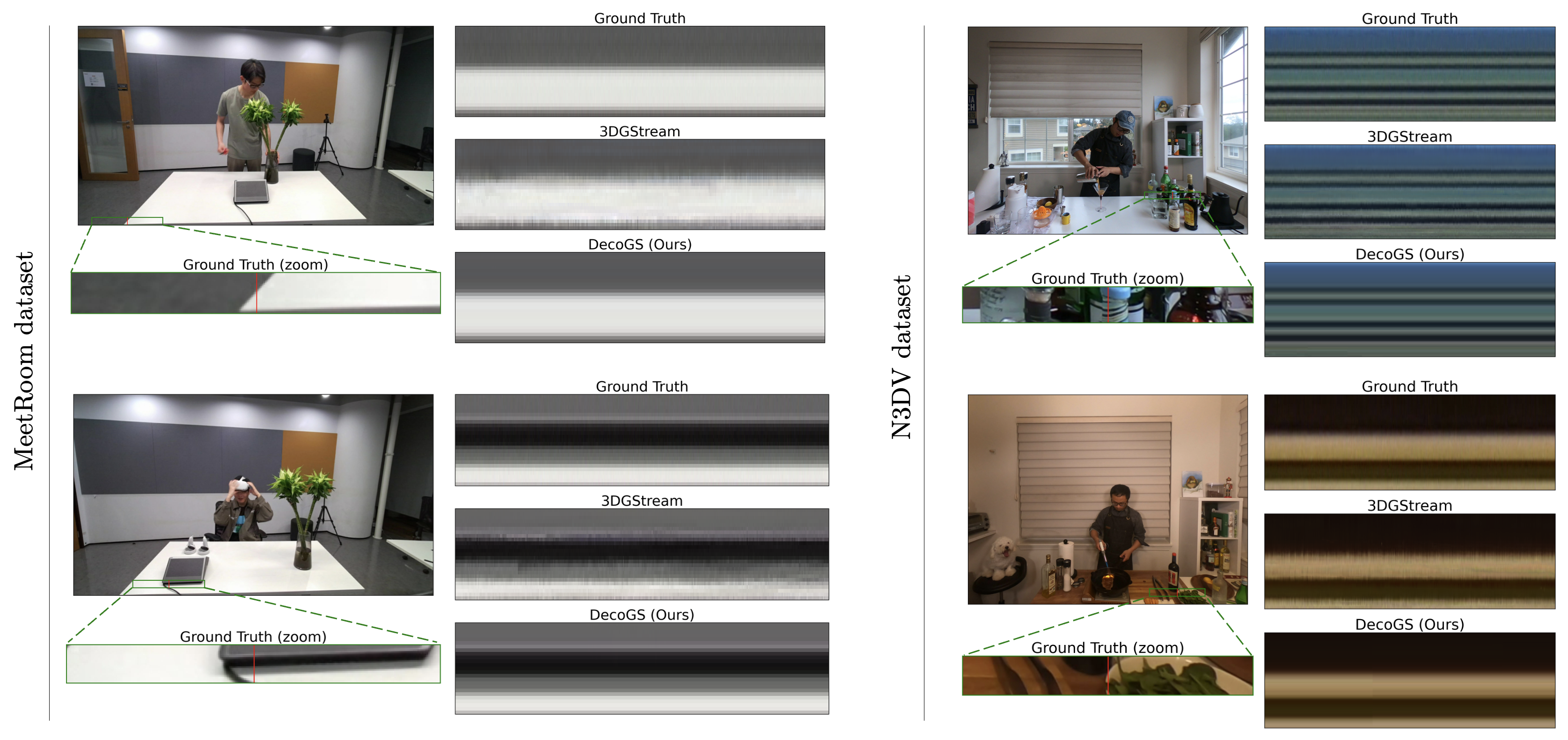}
        \caption{\textbf{Scanlines} across MeetRoom and N3DV datasets. Each image shows a 1 pixel line over the static area across the video.}
        \label{fig:scanline}
    \end{minipage}
    \hfill
    \begin{minipage}[t]{0.41\linewidth}
        \centering
        \includegraphics[width=\linewidth]{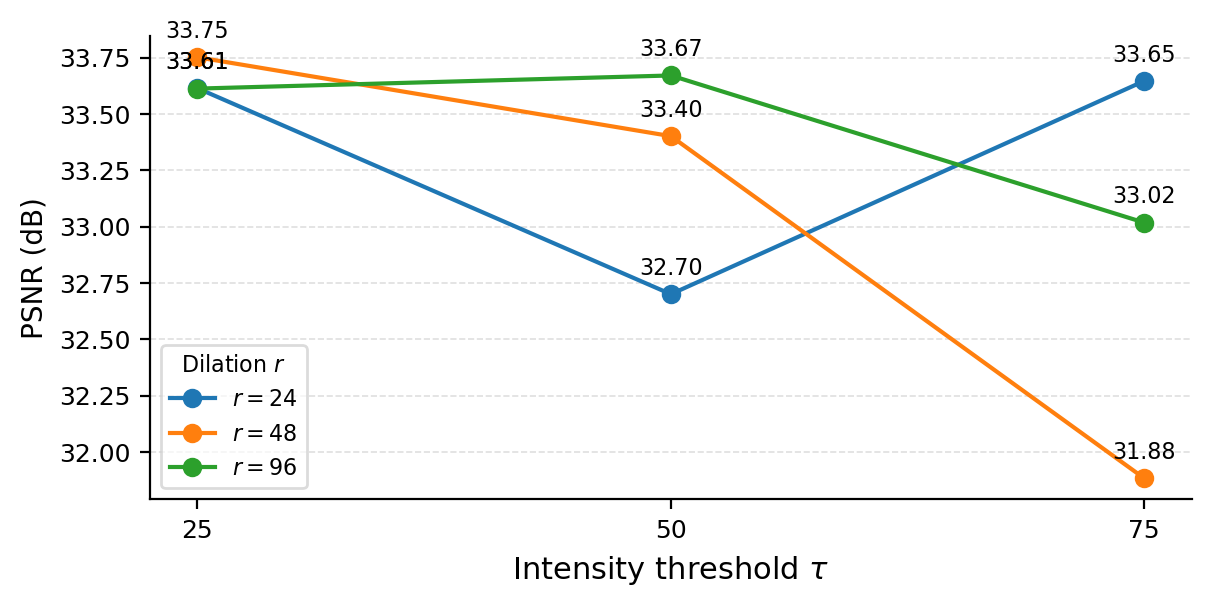}
        \caption{\textbf{Sensitivity of Stage~2} PSNR to intensity threshold $\tau$ and dilation radius $r$ on the discussion scene. DecoGS is robust around $\tau{=}25$, $r{=}48$ (our default).}
        \label{fig:mask_params}
    \end{minipage}
    \vspace{-8pt}
\end{figure*}

The proposed Adaptive Static-Dynamic Decoupling plays a key role in improving spatial clarity: dynamic regions (e.g., moving limbs or objects) are reconstructed with enhanced detail, while static backgrounds remain stable. Moreover, our Selective Gaussian Optimization suppresses redundant updates in non-changing areas, preventing the cumulative artifacts observed in other online methods. As in the “sear steak” and “discussion” scenes, DecoGS maintains fine surface texture and correct object geometry under large inter-frame motion, highlighting its ability to localize and track motion-aware Gaussians effectively. Fig.~\ref{fig:scanline} further illustrates temporal stability through space-time scanlines over static regions: 3DGStream produces a noisy spatiotemporal profile with flickering artifacts~\cite{yun2025compensating}, whereas DecoGS maintains a clean, consistent profile throughout the sequence. Overall, DecoGS achieves photorealistic rendering quality and temporal smoothness, demonstrating clear advantages for real-time FVV streaming.

\subsection{Ablation Study}

\label{sec:ablation}
We conduct an ablation study to analyze the impact of individual components in DecoGS. We evaluate our method across different architectural and training configurations on the N3DV dataset, as shown in \cref{tab:component_analysis}.

\begin{table}[h]
  \centering
  \small
  \renewcommand{\arraystretch}{0.85}
  \begin{tabular}{@{}cccc@{}}
    \toprule
    Select & Focus & Gate & PSNR (dB) $\uparrow$ \\
    \midrule
    --     & --    & --   & 28.61 \\
    \cmark & --    & --   & 30.81 \\
    \cmark & \cmark & --  & 30.83 \\
    \cmark & \cmark & \cmark & 30.92 \\
    \bottomrule
  \end{tabular}
  \caption{\textbf{Component impact} on flame salmon scene of N3DV.}
  \label{tab:component_analysis}

  \smallskip
  \begin{tabular}{lc}
    \toprule
    Setting & PSNR (dB) $\uparrow$ \\
    \midrule
    3DGStream (30 FPS) & 33.39 \\
    \midrule
    No sampling (30 FPS) & 34.40 \\
    Sampling rate = 5 (6 FPS) & 33.68 \\
    Sampling rate = 10 (3 FPS) & 33.25 \\
    \bottomrule
  \end{tabular}
  \caption{\textbf{Effect of temporal sampling} on sear steak of N3DV.}
  \label{tab:fps_analysis}
\end{table}

\para{Dynamic Gaussian Selection.}{ We begin by evaluating the effect of our dynamic 3D Gaussian selection module (\cref{sec:dynamic-gaussian-selection}), which identifies and updates only a sparse set of Gaussians whose 2D projections fall inside view-specific ROIs. Without this selection, the system defaults to dense optimization over all Gaussians, leading to slower updates and degradation of static regions. Enabling selection alone improves PSNR from 28.61dB to 30.81dB, validating that localized updates offer a better tradeoff between adaptation and consistency, and confirming that the key bottleneck in prior methods is indiscriminate optimization, not their choice of motion representation or loss function.}

\para{Focus-Aware Loss. }{We introduce a dynamic region-weighted loss in \cref{sec:focus-aware-loss-func} that increases the contribution of ROI pixels during training. This modification marginally improves the PSNR to 30.83~dB and yields visibly sharper edges in motion regions, where sub-pixel errors are perceptually significant. Qualitative gains are pronounced in motion-rich sequences (e.g., flame in “flame steak”), where standard loss formulations tend to underfit faint boundaries.}

\para{Gradient Gating and Static Suppression.}{When combined with gradient gating in \cref{sec:dynamic-gaussian-selection}, which blocks updates for Gaussians outside of the dynamic region, PSNR further increases to 30.92dB. This configuration enforces strong background stability while focusing learning on changing content. Without gating, we observe subtle ghosting and bleeding in static areas, particularly on walls and stationary surfaces, that accumulate over time. These results confirm the importance of isolating dynamic regions both during selection and optimization.}

\para{Sensitivity to Frame Rate.}{While in pre-recorded captures all frames are available, in real-world streaming scenarios, frame drops are possible. This requires robustness to varying sample rates. To test this, we study the impact of training frame rate (\cref{tab:fps_analysis}). As it decreases from 30~FPS to 6~FPS and 3~FPS, the reconstruction quality degrades from 34.40~dB to 33.68~dB and 33.25~dB respectively. Even at 6~FPS, DecoGS (33.68~dB) outperforms 3DGStream (33.39~dB) on the same scene, while the 3 FPS variant remains on par, demonstrating that DecoGS is robust to sparse frame-rates without significant quality loss.}

\para{Mask Parameter Sensitivity.}{The difference mask is governed by two parameters: intensity threshold $\tau$ and dilation radius $r$. We sweep $\tau$ and $r$ on the ``discussion'' scene and report mean PSNR in \cref{fig:mask_params}. The method is robust around $\tau{=}25$, $r{=}48$, which we adopt as the default. A too-conservative mask ($r{=}24$) under-covers motion boundaries and consistently underperforms, while over-dilation ($r{=}96$) introduces marginal noise. Notably, performance curve is relatively flat across the tested range, confirming that DecoGS requires no per-scene hyperparameter tuning, which is a practical advantage over methods that use scene-specific optical flow or segmentation thresholds.
}

\subsection{Limitations and Future Work}
DecoGS opens several promising avenues for future work. Current change detection can be extended with semantic or foundation model cues to enable intent-aware dynamic region detection that distinguishes transient objects from persistent background without heuristics. Coupling Gaussian representations with language features would further unlock semantic FVV streaming, enabling querying and editing directly in Gaussian space. Finally, all existing online FVV methods, including DecoGS, assume fixed, synchronized, and calibrated cameras; robustness to camera drift and exposure changes remains open for the field. 
\section{Conclusion}
\label{sec:conclusion}

DecoGS demonstrates a fundamental finding: in streaming 3D Gaussian reconstruction, doing less is more. By identifying that fewer than 35\% of Gaussians require updating at any given frame, and building a targeted pipeline around this sparsity, DecoGS achieves state-of-the-art quality with 70× lower temporal flicker, without pretraining, without optical flow, and without increasing per-frame training time. By isolating regions that change over time through fast image-space differencing and updating only a sparse set of 3D Gaussians per frame, DecoGS keeps static regions frozen while preserving temporal coherence and background sharpness. Our focus-aware loss further enhances detail in dynamic areas, and selective densification adapts to scene changes.

Experimental results on multiple real-world dynamic scene datasets demonstrate that DecoGS outperforms prior streaming and offline methods in both quality and efficiency. Notably, our approach achieves higher reconstruction quality while maintaining comparable training and rendering speeds, without large-scale pretraining. More broadly, DecoGS suggests that selective optimization may be a general principle for efficient incremental scene learning, with applicability beyond Gaussian splatting to any representation that accumulates gradient state over time.
\section{Acknowledgments}
\label{sec:acknowledgments}
We thank Justin Solomon and the team of volunteers for organizing the MIT Summer Geometry Initiative (SGI) and for fostering the collaboration that led to this work.

{
    \small
    \bibliographystyle{ieeenat_fullname}
    \bibliography{main}
}

\end{document}